\pdfoutput=1
\documentclass[11pt]{article}

\usepackage{emnlp2021}
\usepackage{times}
\usepackage{latexsym}
\usepackage{natbib}
\usepackage[T1]{fontenc}
\usepackage[utf8]{inputenc}

\usepackage{subfigure}

\usepackage{microtype}
\usepackage{amssymb}

\usepackage{arydshln}
\usepackage{amsmath}
\usepackage{amsthm}
\usepackage{booktabs}
\usepackage{algorithm}
\usepackage{algorithmic}
\usepackage{booktabs}
\usepackage{multirow}
\usepackage{graphics}
\usepackage{ulem}
\usepackage{todonotes}
\usepackage{svg}
\usepackage{xcolor,colortbl}
\usepackage{fixltx2e}
\usepackage{enumitem}

\definecolor{aseem}{cmyk}{0.07, 0.99, 0, 0.00}

\definecolor{usha}{cmyk}{0.55, 0.51, 0.00, 0.14}

\usepackage{soul,xcolor}
\newcommand{\comment}[1]{}

\newcommand{\model}{\texttt{GeM}}

\title{A Computational Approach to Understand Mental Health from Reddit: Knowledge-aware Multitask Learning Framework\thanks{*To appear in International AAAI Conference on Web and Social Media (ICWSM 2022) Copyright © 2022. All rights reserved.}}

\author{Usha Lokala\textsuperscript{1}, Aseem Srivastava\textsuperscript{2}, Triyasha Ghosh Dastidar\textsuperscript{3}, Tanmoy Chakraborty\textsuperscript{2}, Md Shad Akthar\textsuperscript{2}, \\
\textbf{Maryam Panahiazar\textsuperscript{4}, Amit Sheth\textsuperscript{1}}\\\\
\textsuperscript{1}Artificial Intelligence Institute, University of South Carolina, USA \\
\textsuperscript{2}Dept. of CSE, IIIT Delhi, India
\textsuperscript{3}Dept. of CSE, BITS Pilani, India\\
\textsuperscript{4}Bakar Computational Health Sciences Institute, University of California, USA\\
\small{
nlokala@email.sc.edu, \{aseems,shad.akhtar,tanmoy\}@iiitd.ac.in, f20170829@hyderabad.bits-pilani.ac.in, maryam.panahiazar@ucsf.edu, amit@sc.edu}
}

\begin{document}
\maketitle
\begin{abstract}
Analyzing gender is critical to study mental health (MH) support in CVD (cardiovascular disease). The existing studies on using social media for extracting MH symptoms consider symptom detection and tend to ignore user context, disease, or gender. The current study aims to design and evaluate a system to capture how MH symptoms associated with CVD are expressed differently with the gender on social media. We observe that the reliable detection of MH symptoms expressed by persons with heart disease in user posts is challenging because of the co-existence of (dis)similar MH symptoms in one post and due to variation in the description of symptoms based on gender. We collect a corpus of $150k$ items (posts and comments) annotated using the subreddit labels and transfer learning approaches. We propose \model, a novel task-adaptive multi-task learning approach to identify the MH symptoms in CVD patients based on gender. Specifically, we adapt a knowledge-assisted RoBERTa based bi-encoder model to capture CVD-related MH symptoms. Moreover, it enhances the reliability for differentiating the gender language in MH symptoms when compared to the state-of-art language models
\footnote{Resources created as a part of the study will be made available upon request to the corresponding author}. 
Our model achieves high (statistically significant) performance and predicts four labels of MH issues and two gender labels, which outperforms RoBERTa, improving the recall by $2.14$\%  on the symptom identification  task and by $2.55$\% on the gender identification  task.

\end{abstract}

\begin{figure}[!t]
    \centering
    \scalebox{0.48}{
    \includegraphics[width=\textwidth]{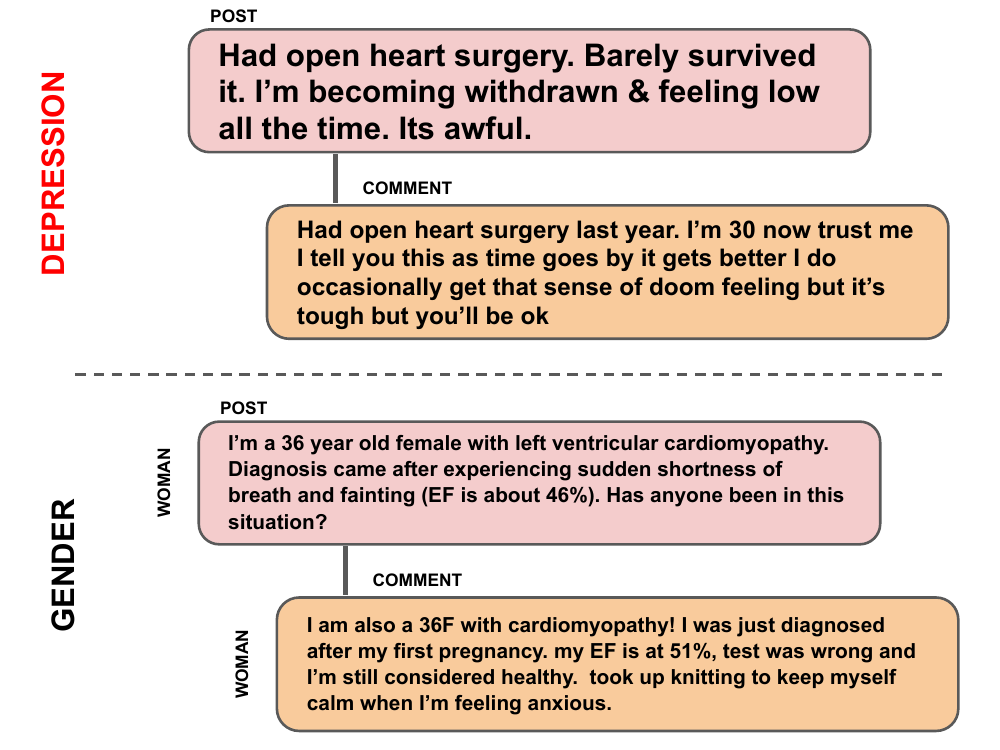}}
    \caption{Our framework of social media posts contains four CVD MH symptoms– Depression, Anxiety, Bipolar, PTSD. We differentiate between gender for these communications. Our computational approach simultaneously identifies these symptoms and the underlying gender.}
    \label{Figure-1}    
    \vspace{-3mm}
\end{figure}

\section{Introduction}
Social media platforms such as Twitter, TalkLife, and Reddit are extremely popular to express someone's feelings and views. In the past few years, people are using them extensively to share their mental and other general health issues. Cardiovascular Disease (CVD) is one such issue and is one of the leading causes of mortality in the United States and globally\footnote{\url{https://tinyurl.com/yh7yde4f}}. Gender, ethnic, racial, and age discrepancies within CVD diagnosis and treatment exist and have been well-reported \cite{kanchi2018gender}. The activation of stress pathways in people leads to the mechanism that links Mental Health (MH) disorders and CVD. 
The American Journal of Medicine study established statistically significant links between MH disorders of depression, anxiety, psychological distress, and CVD \cite{de2018intriguing}. They also found that the link is stronger in the case of depression-related symptoms and it can worsen the symptoms of CVD \cite{daniels2017mental}. An article by the Cleveland Clinic also shows a particularly strong association between depression and CVD\footnote{\url{https://tinyurl.com/n2spyx2j}}. Thus, MH disorders are now established as risk factors for developing CVD, similar to the classic risk factors such as hypertension, diabetes mellitus, substance abuse, etc. 
Another dimension along which CVD can be analyzed is a person's gender. 

Even though the general CVD risk factors in men and women are similar, the most common CVD symptoms and when and how they manifest vary significantly in women compared to men. For example, women are more likely to experience shortness of breath, nausea, backache, and vomiting while that is not the case for men\footnote{\url{https://tinyurl.com/593hx3ae}}. As we see, a lot is known about the individual conditions and conditions in pairs such as CVD and MH, and CVD and gender-specific analysis; however, to the best of our knowledge, CVD has not been studied with both MH and gender-related risk factors. Obtaining clinical data for such a task is necessary to practice safe and ethical standards; however, it is highly cumbersome because it requires several expensive procedures and introduces a considerable time lag in procurement \cite{kilbridge2003cost}. 

  In the digital era, it is well known that plenty of domain-specific knowledge on CVD and various MH disorders from reliable sources exist \cite{Chaddha2016-gf,Gkotsis2016-yr,Ladeiras-Lopes2020-cq}. However, the recent pandemic and overwhelmed healthcare system have pushed people to flock to online community platforms such as Reddit to post their concerns about gender-related issues on subreddits such as \textit{r/askmen} and \textit{r/askwomen}. This is possibly explained by \textbf{``Reddit potentially being the Internet's best support group"\footnote{\url{https://tinyurl.com/3p9xjnmd}}}. Though social media has been employed in past research for analysis of individual cases concerning CVD,  the majority of it is focused more on Facebook, and Twitter \cite{ladeiras2020social}. The anonymity afforded on Reddit causes the posts to be more honest and less distorted by social perception and, thus, of sufficient quality to perform an investigative study. As millions of people use social media platforms for seeking MH support, understanding the symptoms varying with gender is crucial for web-based intervention. However, clinicians and health professionals are more interested in understanding MH disorders connected with the disease \cite{dhar2016depression}. 

\paragraph{Problem Statement:}In this paper, we seek to investigate the gender based MH detection among Reddit users afflicted by CVD. Specifically, we explore the question -- \textit{can gender detection improve the identification of MH in those with heart disease?} We observed that plenty of individuals openly discuss their MH symptoms on Reddit \cite{Gaur2018-gj,Gkotsis2016-yr,Yadav2021-zx,Kursuncu2019-mw,Kursuncu2018-fb}. However, a CVD MH-related post puts forward the issue of how each gender describes the symptom. In this work, we limit this study to two genders (reason being data inadequate for other genders for this particular study on social media), "man" and "woman', since 'man' and 'woman' include people who may or may not have been assigned 'male' or 'female' at birth. The reason these terms are more appropriate for this work is that, given the source of the data (Reddit is solely self-report), there is no way that we can know for each post whether they are man, woman, cis or trans at the time of birth, so it is more appropriate to use man/woman identity categories as opposed to male/female "biological" categorizations. We refer Our framework of social media posts and comments from Reddit (see Figure \ref{Figure-1}) contains four CVD MH symptoms and two gender labels for these communications.

Formally, we define the two tasks as follows:

Let \textit{$S_i = s_{i1},....,s_{it}$} be the Reddit MH text (post/comment) and \textit{$G_i = g_{i1},....,g_{it}$} be the gender text (post/comment).
For each  $S_i$ and $G_i$, two tasks are performed.

\noindent\textbf{Task 1: Symptom Identification (SI) in CVD posts.} The task is to identify the MH symptom, the post is talking about in the context of CVD. For each post/comment $S_i$ in our dataset, we identify one of the four symptoms -- \textit{depression}, \textit{anxiety}, \textit{bipolar}, or \textit{PTSD}. 

\noindent\textbf{Task 2: Gender Identification (GI) in CVD posts.} In this task, we aim to detect the gender (man/woman) of the person showing CVD-related MH symptoms. This is to note that the post may not always be written by the user himself/herself; instead, in some cases, the post can be written by a close relative or a friend as well. For example, the following post, \textit{She needs a cardiac procedure, which she is refusing to have. The cardiologist told her she can guarantee her she'll have a heart attack without it but still insists she's not having the procedure.} is originally posted by a \textit{man}. With so many woman-specific keywords, we want our model to predict gender correctly irrespective of the gender masking and writer of the post.

\textbf{{Proposed Method and Contributions:}} We compile a high-quality CVD dataset (CVDS) by crawling six subreddits. The CVDS dataset consists of a total of $16k$ posts and comments extracted from the crawled $150k$ Reddit data points. 
Moreover, the annotated dataset is generated using the automatic labelling methods and is validated by the   domain experts. We develop a computational approach to understand the gender-based MH symptoms on social media. Our approach, which is knowledge aware, task adaptive RoBERTa based bi-encoder model, simultaneously identifies these symptoms and the underlying gender. We summarize our contributions below:
\begin{enumerate}[leftmargin=*]
\setlength{\itemsep}{0em}
  \item We propose a multi-task learning (MTL) framework for identifying MH symptoms that consider gender language fused in the CVD-related user posts. This is the first study in the CVD domain that considers gender language in MH-related posts to the best of our knowledge.
  \item To justify the variations in the gender language description of symptoms, we propose a \textbf{G}ender \textbf{E}nabled \textbf{M}H framework, \model, which is a RoBERTa-based bi-encoder model. The performance of \model\ is improved for the task of SI along with the task of GI in a MTL paradigm.
  \item To evaluate our model, we created a corpus of $150k$ data points with $16k$ posts and comments annotated (CVDS) with four MH symptoms. Also, these posts were labeled with the gender classes -- man and woman.
\end{enumerate}

\section{Related Work}

\paragraph{MH Setting:}
Due to the superior performance of deep learning algorithms in data-rich healthcare applications \cite{miotto2018deep}, computational methods to model MH outcomes using deep learning are fast gaining prominence. \citet{shatte2019machine} explored the application of machine learning techniques in MH for diagnosis, prognosis, and treatment. 

\citet{pruksachatkun2019moments} used machine learning to identify changes in MH behavior using topic and sentiment analysis of the posts. \citet{barney2011explicit}  modeled depression concerns of people. \citet{chancellor2020methods} interpreted MH based on social media knowledge. \citet{gaur2021can,gaur2021characterization} identified MH needs of people posting on online Reddit communities using deep learning methods based on time-variant and time-invariant analyses. The recently successful Transformer architecture has also been employed to analyze MH using conversational data from social media apps and online domain knowledge. \citet{dolbir2021nlp} used Transformer models such as GPT-2 to understand the MH condition of the user post. As we can see, there is a large body of work in utilizing deep learning for MH prediction and analysis \cite{akbulut2017gender}.  

\paragraph{CVD Setting:} \citet{mathur2020artificial} surveyed techniques  to predict CVD outcomes, risk factors, early rule-based symptoms limited the use of computational technology for clinical decision-making. Thus, following the advent of deep learning, a wide range of CVD applications, including prediction of drug therapy, pharmacogenomics, heart failure management, cardiovascular imaging, and diagnostics, has been developed \cite{kalinin2018deep, li2020prediction,powles2017google}. Social media analysis on prediction of CVD-related symptoms, risk factors, behaviors and outcomes, to our knowledge has not been studied before. 

\paragraph{Gender Detection:} Gender detection from various types of data including images and text on social media such as Facebook, Twitter and even from different language and dialects has been extensively studied using deep learning models \cite{elsayed2020gender}. \citet{akbulut2017gender} utilized knowledge from local receptive fields and CNNs to identify the gender of a person from their faces. \citet{kowsari2020gender} used an ensemble method   to identify the gender from social networks. \citet{elsayed2020gender} identified gender from Egyptian Arabic dialect tweets using a variety of neural networks, including vanilla deep networks, CNNs, LSTM, GRU, both uni-directional and bi-directional. However, there is little work on gender detection from a resource-rich platform such as Reddit. Thus, we see that Reddit has been relatively less explored for CVD and gender detection even if MH research has been carried out on social media platform. Thus analysis of Reddit for a cross-sectional study of these MH symptoms and gender factors is a highly challenging and necessary research problem.Furthermore, to the best of our knowledge, though deep learning architectures have been developed in healthcare for CVD, MH, gender-based studies as well as multimodal data, no prior work has investigated a combination of all three factors. 

\begin{table}[t]
\centering
\resizebox{\columnwidth}{!}
{
\begin{tabular}{c c c c c c}
\hline
\textbf{  } & \textbf{  } & \textbf{Posts} & \textbf{Comments} & \textbf{Users (P)} & \textbf{Users(C)} \\
    \hline
    Depression &  & 2125 & 2281 & 1966 & 1911 \\
    Anxiety &  & 1983 & 2498 & 1635 & 1804 \\
    Bipolar &  & 1796 & 2016 & 1044 & 1128 \\
    PTSD  & & 1831 & 2027 & 1155 & 969 \\
    \hline 
    man & & 3554 & 4637 & 2645 & 3225 \\
    woman & & 4181 & 4185 & 3014 & 2358 \\
  
\end{tabular}}
\caption{Dataset statistics on the CVDS corpus used in our experiments for SI and GI tasks. (P) and (C) represent the set of users who have posted and commented, respectively.}
\label{tab:dataset:stats}
\end{table}

\section{The CVDS Dataset Development}
In this section, we describe our effort in developing the CVD dataset using Reddit. We also describe our annotation procedure for the same. We use the entire dataset for task adaptive pre-training and annotate a subset of them containing $8k$ posts and $8k$ comments.
\begin{table}[t]
\centering
\scalebox{0.85}{

\begin{tabular}{c c c c c c}
\hline
\textbf{  } & \textbf{  } & \textbf{Posts} & \textbf{Comments} & \textbf{Users (P+C)}\\
    \hline
    r/AskMen &  & 13659 & 22706 & 8347 \\
    r/AskWomen &  & 14823 & 21673 &7955 \\
    
\end{tabular}}
\caption{Dataset statistics on the gender knowledge base corpus used in transfer learning to annotate data with gender labels. (P) and (C) represent the set of users who have posted and commented, respectively.}\label{tab:dataset:gender}
\end{table}

\paragraph{Data Collection:}

We collect posts and comments from MH-related subreddits for this study. We consider four MH subreddits  -- r/depression, r/anxiety, r/PTSD, and r/bipolar. We choose only these subreddits as these are the major CVD symptoms related to MH as suggested in \cite{Cdc2021-eu,ORMEL2007325}. 
Apart from this, we collect data from two gender-related subreddits -- r/AskMen and r/AskWomen. The Reddit API was used to crawl six subreddits for a total of $150k$ posts and comments.

\paragraph{Identifying CVD Discussions:}
We crawl the data using a carefully curated CVD lexicon consisting of around 250 CVD-related terms obtained from various sources like SNOMED-CT\footnote{\url{https://tinyurl.com/nczubnzr}}, DataMed\footnote{\url{https://datamed.org/}} and  ICD-10\footnote{\url{https://tinyurl.com/kyky45kz}}. Moreover, we append the list with commonly used CVD-related topics and abbreviations such as CABG (Coronary artery bypass graft), CAD (Coronary artery disease), CHD (Coronary heart disease), and CHF (Congestive heart failure). We collect these CVD topics and abbreviations from Texas Heart Institute \cite{Texasheart_2017-rw}. 
Utilizing the compiled list, we obtain $\sim16K$ posts and comments from $5,400$ users. The CVDS data stats are presented in Table \ref{tab:dataset:stats} with number of posts and comments for each MH symptom and Gender and number of users who posted and commented.

\paragraph{Automatic Labelling and Annotations:}
The posts in these subreddits are assigned a class label corresponding to the name of the MH disorder they are associated with. We consider posts with a number of upvotes of more than 10 and a minimum token length of 50 to maintain the quality. We further remove any URLs or usernames that could potentially contain sensitive information.

For generating gender labels, we implement an inductive transfer learning approach with BERT \cite{mozafari2019bert}.
For this, we additionally crawl $72k$ posts from r/AskMen and r/AskWomen. We observed that $92$\% of the posts posted in r/askmen are posted by man and the $94$\% of posts from r/askwomen are posted by woman. We split this dataset into train, dev, and test sets (75:5:20). We train a BERT-based text classifier for 10 epochs using learning rate of 1e-5, batch size of 32 to distinguish the posts written by man and woman. Subsequently, gender labels of $16k$ posts are fetched in the CVDS dataset. We also examine manual inter-annotator agreement among  for gender and symptom labels of 300 posts and comments which is 0.72 with biomedical informatics specialist co-author to validate the annotations. The manual annotations are evaluated in same way as automated labels and our best F measure against ground truth is 0.70. The stats of the posts and comments used for training the transfer learning model are found in Table \ref{tab:dataset:gender}.   

The results for transfer learning using BERT are reported in Table \ref{Transfer Lear}. 

\begin{table}[t]
\centering
  {
  \begin{tabular}{lccc}
    \toprule[1pt]
        \bf TL-BERT Model & \bf Precision & \bf Recall &\bf F1 \\
        \toprule[1pt]
            Gender Post & $87.36$ & $83.47$ & $85.92$ \\   
            Gender Comment & $78.78$ & $79.32$ & $78.90$ \\ \bottomrule
    \end{tabular}
  }
\caption{Validation results for GI through the transfer learning approach on generic (non-CVD) related posts. The trained model is then used to obtain the gender labels for the posts and comments in the CVDS dataset.}
\label{Transfer Lear}
\end{table}

\paragraph{Anonymization and Data Statistics:}
The statistics of the CVDS and gender datasets are shown in Tables \ref{tab:dataset:stats} and  \ref{tab:dataset:gender}, respectively.

An important aspect that we need to consider while working with health-related issues is to respect the users' privacy and adhere to good ethical practices adapted by previous research \cite{Henderson2013-nk,Alim2014-bx}. Therefore, similar to \newcite{Matthews2017-cn}, we censor several sensitive information, such as user names, personal information, platform identifiers, and urls which might be directly linked to the users' identity, from the collected posts and comments. 
We also acknowledge that the current work does not make any clinical diagnosis or treatment suggestions in any manner whatsoever. 

\begin{figure*}[!htbp]
    \centering
    \scalebox{1}{
    \includegraphics[width=\textwidth]{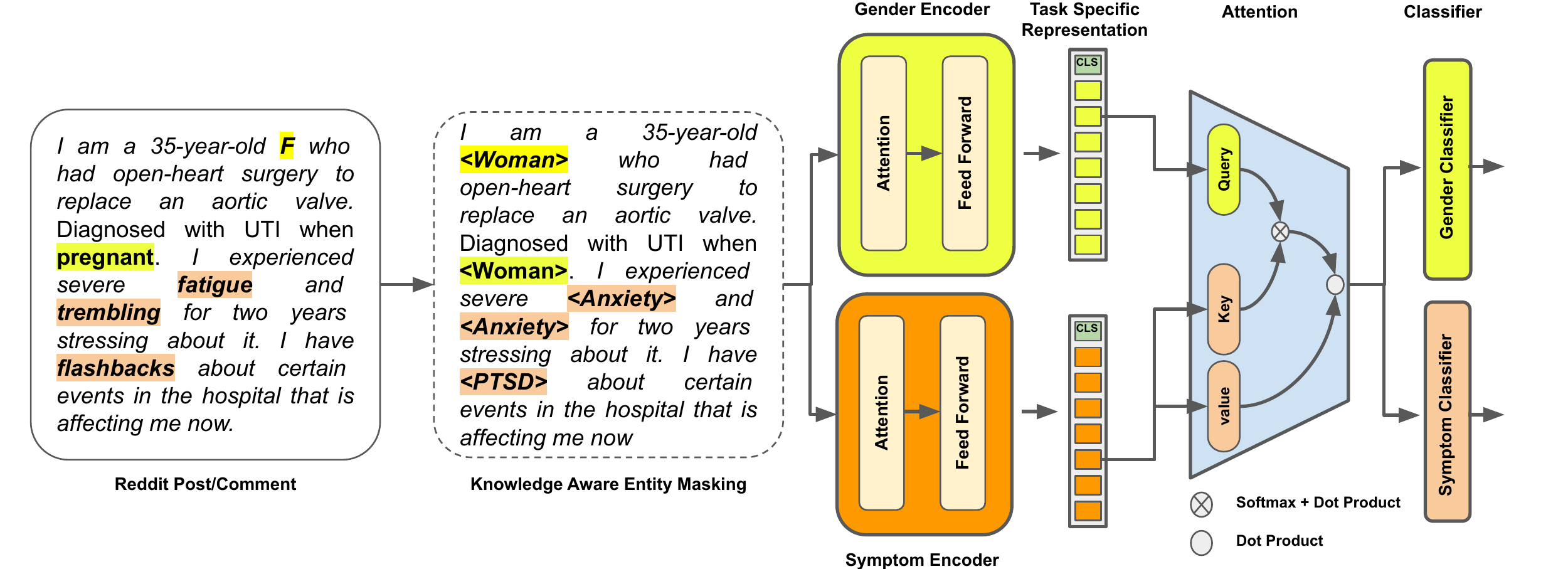}}
    \caption{Proposed model architecture of \textbf{\model}\ for SI and GI tasks. }
   
    \label{fig:arch}
 
\end{figure*}

\section{Proposed Methodology}
We develop a computational approach, \model\ as shown in  Figure \ref{fig:arch} to understand the gender-specific MH symptoms in CVD. We formulate \model\ as a model consisting of knowledge aware entity masking combined with the task adaptive RoBERTa model. Here, we describe the drug abuse ontology as one of the sources of knowledge and the other modules in our architecture.

\paragraph{Drug Abuse Ontology:} 
The Drug Abuse Ontology (DAO) \cite{cameron2013predose, Lokala2020-ay} is a domain-specific knowledge source containing drug and health-related classes, properties, relationships, and instances. Apart from medical terms, it includes concepts of MH disorders and symptoms aligned with the DSM-5\footnote{Diagnostic and Statistical
Manual of Mental Disorders - 5th Edition} scale. Initially, it is used to determine user knowledge, attitudes, and behaviors related to non-medical use of buprenorphine and other illicit opioids through analysis of web forum data. Later, this ontology evolved to understand MH disorders and trends of cannabis product use in the context of changing cannabis legalization policies in the USA. Recently, this proved effective in capturing gleaning trends in the availability of novel synthetic opioids through analysis of crypto market data. In this study, we leverage DAO to map the symptoms we find on social media to the parent concepts for MH conditions and associated symptoms. For example, properties for `\textit{anxiety}' include terms such as `feeling jittery', `antsy', `feeling doomed', `nausea', `lack of concentration', and `throwing up'.  

\paragraph{Entity Masking with External Knowledge:}

\textit{Symptom Masking}: Extracting entities and relationships related to MH symptoms and gender-like concepts/synonyms/slang terms requires domain knowledge. For a text $S_i$, the concepts/symptoms/synonyms related to MH labels (depression, anxiety, PTSD, bipolar) are masked using the MedDRA Ontology\footnote{\url{https://tinyurl.com/s2upzusc}} and DAO. We use these ontologies to map entities and n-grams in a post/comment to their parent concept in the ontology. We further leverage the potential of detecting clinical symptoms in post/comment as per the clinical questionnaires like PHQ-9 \cite{kroenke2001phq} for Depression, GAD-7 \cite{jordan2017psychometric} for Anxiety, CAPS-5 \cite{weathers2018clinician} for PTSD, and BDRS \cite{berk2007bipolar} for Bipolar to perform masking \cite{yazdavar2017semi, Yadav2021-zx}. This approach provides the model with more context about the main entities of interest in the text. It removes noise in the data, thus contributing to the robustness of the model without overfitting that may happen if the model tries to learn noise in data. Our ablation study (Table \ref{ablation}) shows that external knowledge adds value to data in the model decision-making. The reason for masking is not to bias the model to specific entities. For example, the entities or symptoms related to PTSD  in the post and comment are masked with <PTSD>; similarly, entities related to anxiety are masked with <anxiety>, and so on for the other three labels. 

\textit{Gender Masking}:
We observe that the terms like \textit{`F'} and \textit{`M'} on subreddits indicate woman and man, respectively. Similarly, there are many other terms that indicate gender indirectly. For gender word masking, we use `Gendered words'\footnote{\url{https://github.com/ecmonsen/gendered_words}} lexicon which is created from the WordNet synset. Each word in this dataset is manually tagged as gender-neutral (n), man or masculine (m), woman or feminine (f), or other (o). However, for this study, we only use `man' and `woman' tags. Words that cannot be extracted from WordNet are manually added, including personal (he, she, him, her) and possessive (his, hers, her) pronouns. The words related to gender in the post/comments are masked with either <man> or <woman>. For example, bachelorette is masked as <woman> and bachelor is masked as <man>. The input to the model is a Reddit post or comment, and the output is the MH symptom and gender.

\paragraph{Knowledge Assisted Bi-encoder Model:}
We propose a knowledge assisted bi-encoder model based on RoBERTa \cite{Liu2019-wq} for identifying symptom and gender as shown in Figure \ref{fig:arch}. We multitask over the two tasks of SI and GI with data in CVDS. We observe that bi-encoder architecture with external knowledge and attention provide more context than working with the encoder itself. The two encoders Symptom-encoder (S-encoder) and Gender-encoder (G-encoder) for symptom and gender, respectively, are pre-trained on RoBERTa. The S-encoder is for understanding the context of MH symptoms, whereas, the G-encoder is for understanding the gender-related language in the post/comment. 

\paragraph{Task Adaptive Pretraining:}
Both S-encoder and G-encoder are initialized using the weights learned by 
RoBERTa\textsubscript{BASE}. We further perform a task-adaptive pre-training \cite{Gururangan2020-iv} of the two encoders to adapt to the symptom and gender-specific context. For this additional pre-training of the two encoders, we use the datasets of $65k$ posts and $150k$ comments from several MH subreddits (bipolar, anxiety, depression, and PTSD) and gender-related subreddits (r/AskMen, r/AskWomen). For this task, the masked language modeling task for pre-training (10 epochs, batch size 32) is used.

\paragraph{Attention Layer:}
Subsequently, we compute attention over the two encodings for generating MH context-aware representation of gender text. Our query is the gender text encoding $e_i^{(G)}$, while the key and value are the symptom text encoding $e_i^{(S)}$. We compute the attention as:
\begin{equation}
 a_i(e_i^{(S)},e_i^{(G)}) = softmax(\frac{e_i^{(S)}e_i^{(G)}}{\sqrt{d}})e_i^{(G)} \nonumber
\end{equation}
where $d = 768$ is the hidden dimension in RoBERTa\textsubscript{BASE}. Next, we add the encoded gender post $e_i^{(G)}$ with its representation transformed through attention $a_i(e_i^{(S)}, e_i^{(G)})$ to extract a residual mapping, $h_i^{(G)}$ \cite{He2016-ry}. It forms the final MH context-aware representation of the input post.

\paragraph{Prediction Layer:}

Finally, we take the final representations of the [CLS] tokens and pass it through two separate linear layers for the gender and symptoms predictions. 

\section{Experiments and Results} 
In this section, we present our experimental results in identifying MH symptoms with underlying gender. We compare the performance of \model\ against state-of-the-art models used in similar tasks. Moreover, we also present our prediction analysis and how challenging it is to predict MH symptoms based on gender.

\paragraph{Baseline:}
(1) LSTM \cite{hochreiter1997long};
(2) BERT \cite{Devlin2018-co}: This is a BERT base model which is finetuned for individual tasks as single-task learning (STL) for MH and gender detection. We also implement BERT in the MTL setting for the tasks of SI and GI.
(3) GPT-2 \cite{Radford2019-om}. 
(4) RoBERTa\textsubscript{BASE} \cite{Liu2019-wq}. 

\paragraph{Experimental Setup:}
For experiments, we split the CVDS dataset into 75:5:20 ratio for the train, development, and test sets, respectively. We fine tune the hyper-parameters using the development set. Each model is trained for 10 epochs with a learning rate of 1e-5, batch size of 64. We use cross-entropy loss and ADAM \cite{kingma2014adam} for the optimization. For regularization, we use dropout \cite{hinton2012improving} with a probability of 0.2. For both tasks, we report recall, precision, and F1-score. 

\begin{table}[t]
\centering
  \scalebox{0.64}{
  \begin{tabular}{ccccccc}
    \toprule[1pt]
        \multirow{2}{*}{\bf Model} &  
        \multicolumn{3}{c}{\bf SI-Posts} & 
        \multicolumn{3}{c}{\bf SI-Comment} \\
        & \bf Precision & \bf Recall & \bf F1 & \bf Precision & \bf Recall & \bf F1 \\
        \toprule[1pt]
            LSTM\textsubscript{STL} & $71.67$ & $70.15$ & $70.90$ & $75.51$ & $75.53$ & $75.51$ \\
            BERT\textsubscript{STL} & $76.35$ & $76.33$ & $76.33$ & $76.53$ & $76.33$ & $76.42$ \\
            GPT-2\textsubscript{STL} & $76.23$ & $76.85$ & $76.53$ & $76.79$ & $76.60$ & $76.69$ \\
            RoBERTa\textsubscript{STL} & $82.73$ & $82.41$ & $82.56$ & $80.10$ & $81.28$ & $80.68$ \\
            BERT\textsubscript{MTL} & $75.24$ & $77.12$ & $76.16$ & $74.65$ & $77.45$ & $76.53$ \\
            RoBERTa\textsubscript{MTL} & $81.34$ & $83.15$ & $82.23$ & $81.25$ & $83.74$ & $82.47$ \\
            \bottomrule
            \bf \model & $\textbf{84.18}$ & $\textbf{85.29}$ & $\textbf{84.73}$ & $\textbf{83.57}$ & $\textbf{84.07}$ & $\textbf{83.81}$ \\
            \bottomrule
    \end{tabular}}
    \caption{Performance comparison our proposed model with the baselines for SI on both the posts and comments. SI: Symptom Identification.}
    \label{tab:results:SI}
\end{table}

\paragraph{Results:}
Tables \ref{tab:results:SI} and \ref{tab:results:GI} presents the results of \model\ with existing models, indicating that \model\ significantly outperformed the previous architectures.
Tables \ref{tab:results:SI} and \ref{tab:results:GI} report the performance for the SI and GI tasks, respectively. Table \ref{tab:results:class} presents the class-wise results of \model\ for symptom and gender on posts and comments. LSTM performs the worst among the four baselines while RoBERTa achieves the best performance. However, \model\ outperforms RoBERTa with substantial gains demonstrating the performance of our `knowledge aware bi-encoder' model  over other baselines for this task. Basically, we train our baselines in two different ways -- STL and MTL. We can see in Table \ref{tab:results:SI} and Table \ref{tab:results:GI} that MTL framework performs well in the main task (MH symptom detection) and also the secondary task (gender detection). However, BERT and GPT-2 perform similarly in the STL setting while RoBERTa performed better both for for main task  in the STL setting and in MTL setting. But for the gender detection task, BERT performs better than GPT-2 in the STL setting while RoBERTa performs similarly in STL and MTL settings. This demonstrates that providing an additional attention mechanism improves the performance of \model\ over the strong self-attention mechanism of RoBERTa for this task. To get a deeper understanding of how \model\ performed over the baseline models, we observe the following on the classification of posts and comments on SI and GI tasks.

\begin{itemize}[leftmargin=*]
\setlength{\itemsep}{0pt}
    \item \textbf{Understanding MH symptom:} Table \ref{tab:QA} shows the capability of our model to capture symptoms like depression, anxiety, bipolar and PTSD. Our model performs better than GPT-2 and LSTM in predicting Depression and PTSD. The reason why BERT misclassified Bipolar and Anxiety maybe because of BERT model trained on sources like Wikipedia, Books compared to MH symptoms described in social media. \model\ is able to detect classes-bipolar and PTSD with high accuracy among posts and comments compared to classes-depression and anxiety (Table \ref{tab:results:class}). Even
with just the posts, \model\ is able to predict the Biploar class with an F1 score of 87\%.

    \item \textbf{Understanding Gender:} For the posts where gender was not explicitly mentioned, most of the baselines are prone to a false classification as shown in Table \ref{tab:QA}. BERT fails to model the meaning of gender context inside post. If the post talks about his/her family, some of the baselines fail to understand the primary gender the post talks about. However, our model with the MH aware attention information sharing unit tends to have better coverage for identifying the symptoms related to CVD in predicting the primary gender the post is meant to talk about. \model\ can detect gender classes with high accuracy among both posts and comments relatively well (Table \ref{tab:results:class}). 
\end{itemize}

\begin{table}[t]
\centering
  \scalebox{0.64}{
  \begin{tabular}{ccccccc}
    \toprule[1pt]
        \multirow{2}{*}{\bf Model} &  
        \multicolumn{3}{c}{\bf GI-Posts} & 
        \multicolumn{3}{c}{\bf GI-Comment} \\
        & \bf Precision & \bf Recall & \bf F1 & \bf Precision & \bf Recall & \bf F1 \\
        \toprule[1pt]
            LSTM\textsubscript{STL} & $64.95$ & $65.11$ & $65.02$ & $70.99$ & $70.78$ & $70.88$ \\
            BERT\textsubscript{STL} & $78.78$ & $79.32$ & $79.04$ & $76.53$ & $76.33$ & $76.42$ \\
            GPT-2\textsubscript{STL} & $70.20$ & $70.75$ & $70.47$ & $72.57$ & $71.27$ & $71.91$ \\
            RoBERTa\textsubscript{STL} & $78.80$ & $78.34$ & $78.56$ & $80.53$ & $79.42$ & $79.97$ \\
            BERT\textsubscript{MTL} & $77.65$ & $75.34$ & $76.47$ & $77.54$ & $74.38$ & $75.92$ \\
            RoBERTa\textsubscript{MTL} & $79.32$ & $77.32$ & $78.30$ & $79.47$ & $77.65$ & $78.54$ \\
            \bottomrule
            \bf \model & $\textbf{80.34}$ & $\textbf{79.87}$ & $\textbf{80.10}$ & $\textbf{80.53}$ & $\textbf{79.67}$ & $\textbf{80.09}$ \\
            \bottomrule
    \end{tabular}
  }
\caption{Performance comparison our proposed model with the baselines for GI on both the posts and comments. GI: Gender identification.}
\label{tab:results:GI}
\end{table}

\begin{table}[t]
\centering
  \scalebox{0.64}{
  \begin{tabular}{ccccccc}
    \toprule[1pt]
        \multirow{2}{*}{\bf Class} &  
        \multicolumn{3}{c}{\bf Posts} & 
        \multicolumn{3}{c}{\bf Comments} \\
        & \bf Precision & \bf Recall & \bf F1 & \bf Precision & \bf Recall & \bf F1 \\
        \toprule[1pt]
            Depression & $82.71$ & $84.24$ & $83.46$ & $78.45$ & $80.63$ & $79.52$ \\
            Anxiety & $80.28$ & $84.59$ & $82.37$ & $79.26$ & $81.81$ & $80.51$ \\
            Bipolar & $88.87$ & $85.36$ & $87.07$ & $87.28$ & $87.46$ & $87.36$ \\
            PTSD & $84.66$ & $86.97$ & $85.79$ & $89.26$ & $86.38$ & $87.79$ \\
            man & $79.37$ & $77.26$ & $78.30$ & $80.78$ & $76.15$ & $78.39$ \\
            woman & $81.32$ & $82.51$ & $81.91$ & $80.29$ & $83.18$ & $81.70$ \\
            \bottomrule
         
    \end{tabular}}
    \caption{Class-wise performance of \model\ for MH symptoms and gender on both the posts and comments.}
    \label{tab:results:class}
\end{table}

\begin{table*}[t]
    
    \centering
    \resizebox{\textwidth}{!}
    {
    \begin{tabular}{p{26em} |c| c| c ccc}       
    \hline
    
\multirow{2}{*}{\textbf{Post/Comment}} & \multirow{2}{*}{\textbf{Case}} & \multirow{2}{*}{\textbf{Actual}} & 
\multicolumn{4}{c}{\textbf{Predicted}}  \\ \cline{4-7} 
 & & & \textbf{BERT\textsubscript{MTL}} & \textbf{GPT-2} & \textbf{RoBERTa\textsubscript{MTL}} & \textbf{GeM} \\ \hline


         \hline
         \textit{I'm having heart surgery in 2 weeks. This is too much for someone who has been suffering from panic attacks} & \multirow{4}{*}{Symptom} & \multirow{2}{*}{Anxiety} & \multirow{2}{*}{\textcolor{red}{PTSD}} & \multirow{2}{*}{\textcolor{red}{Bipolar}} & \multirow{2}{*}{Anxiety} & \multirow{2}{*}{Anxiety} \\ \cdashline{1-1} \cdashline{3-7} 
         \textit{I've noticed that a lot of my CVD symptoms have escalated over time. My flash backs have become much more 'vivid', the adrenaline response to panic has increased, my libido has plummeted.} & & \multirow{4}{*}{PTSD} & \multirow{4}{*}{PTSD} & \multirow{4}{*}{\textcolor{red}{Depression}} & \multirow{4}{*}{\textcolor{red}{Anxiety}} & \multirow{4}{*}{PTSD} \\ \hline
         
         \textit{I was on Prozac for a while, but I was still having panic attacks once or twice a month. I think mine are related to my menstrual hormones} & \multirow{4}{*}{Gender} & \multirow{3}{*}{woman} & \multirow{3}{*}{\textcolor{red}{man}} & \multirow{3}{*}{woman} & \multirow{3}{*}{\textcolor{red}{man}} & \multirow{3}{*}{woman} \\ \cdashline{1-1} \cdashline{3-7}
         \textit{My grandma had a stress-induced heart attack 4 years ago. My grandpa, her husband of over 60 years, had dementia, and she was his sole carer} & & \multirow{3}{*}{woman} & \multirow{3}{*}{\textcolor{red}{man}} & \multirow{3}{*}{\textcolor{red}{man}} & \multirow{3}{*}{woman} & \multirow{3}{*}{woman} \\
         \hline
    \end{tabular}
    }
    \caption{Qualitative analysis of proposed approach GEM, with the baseline models.}
    \label{tab:QA}
\end{table*}

\begin{table*}[t]
    \centering
    \resizebox{\textwidth}{!}
    {

    \begin{tabular}{p{40em} |c|c| c}       
    \hline

        {\bf Post/Comment} & {\bf Error Type} & {\bf Actual} & {\bf Predicted} \\ 
       
         \hline
         \textit{Issues with IUD in uterus since the first heart attack. My grandfather didn't get stronger too after he survived his heart attack. It left him weak and he spent the rest of his days wishing he died. if anything it's left me with a mountain of debt and a deep depression.} & \multirow{4}{4em}{\centering Ambiguity in Gender} & \multirow{4}{*}{woman} & \multirow{4}{*}{\textcolor{red}{man}} \\
         \hline
         \textit{I get panic attacks and depression. The first time I had a panic attack I was driving on the highway and thought that I was going to seriously have a heart attack and die. It was mixture of fear  and panic in one.} & \multirow{3}{5em}{\centering Co-existing Symptom} & \multirow{3}{*}{Anxiety} & \multirow{3}{*}{\textcolor{red}{Depression}} \\
         \hline
         \textit{My ptsd came about from a traumatic experience of terror but my depression and anxiety plummeted. They all experienced high anxiety problems too. The other two ended up with serious heart problems that forced them into surgery and unemployment. We all ended up poor. I ended up homeless.} & \multirow{4}{5em}{\centering Class-level Ambiguity} & \multirow{4}{*}{PTSD} & \multirow{4}{*}{\textcolor{red}{Anxiety}} \\
         \hline
    \end{tabular}
    }
    \caption{Examples showing major errors made by our proposed approach.}
    \label{tab:error}
\end{table*}

\subsection{Error Analysis}
We qualitatively analyze the sources of our errors (Table \ref{tab:error}). The following are the major errors made by our approach.
\begin{enumerate}[leftmargin=*]
\setlength{\itemsep}{0pt}    
    \item \textbf{Co-existing Symptoms:} \model\ is not able to predict correctly if multiple symptoms co-exist, e.g., when depression might often occur with another MH symptom in a post (c.f. Table \ref{tab:error}).
    
    \item \textbf{Class-level Ambiguity:} Regarding the class-wise performance among MH symptoms, the best performing classes are PTSD and bipolar, while the worst performing classes are depression and anxiety (Table \ref{tab:results:class}). The reason can be attributed to the word depression occurs in 23\% of total posts related to other symptoms, and anxiety occurs in 31\% of total posts related to other symptoms. This implies that the model cannot capture such class mentions as it does with other MH symptom classes. This may explain the relatively lower precision for depression and anxiety.
    
    \item \textbf{Ambiguity in Gender:} Posts mentioning about the persons themselves are easy to classify in gender than a post talking about others and then transitioning back to self-description (Table \ref{tab:error}). However, mentions in posts such as `my mother' or `my brother' are correctly predicted in gender as shown in Table \ref{tab:QA}.

\end{enumerate}

\subsection{Ablation Study}
To determine statistical significance, we perform Wilcoxon Signed Rank Test \cite{woolson2007wilcoxon} on MH symptom expressed in posts and comments between where gender information is present and absent. We observe that the cosine similarity between MH symptom expressed by post/comment with gender present is significantly ($p$ < 0.001) lower than  a  gender absent  post/comment. We next analyze the components and training strategies in our approach through an ablation study by removing one component from \model\ and evaluating the performance. To remove the attention component from the model, instead of using attention, we concatenate the symptom post encoding e\textsubscript{i}\textsuperscript{(S)} with the Gender post encoding e\textsubscript{i}\textsuperscript{(G)} and use the concatenated representation as input to the linear layer. For removing the entity masking component, we train our encoders with raw data directly extracted from the social media. Task adaptive pre-training is removed from the model by initializing the model weights from RoBERTa. We present the results on SI task for posts in Table \ref{ablation}. Our major gains come from using entity masking which significantly benefits the SI task (+2.96 precision, +3.16 recall). The result is significantly better than model without entity masking ($p$ < 0.05) for Wilcoxon Signed Rank test which shows that contextualized representation is highly desirable for the SI task in this study. Also, using task adaptive pre-training only leads to minor performance improvement. We further observe that attention improved the precision of the model by $3.16$\% making every aspect of the model contributing to the performance of this task. 
\begin{table}[!t]
\resizebox{1\linewidth}{!}{
\begin{tabular}{l|l|l|l}
\hline
\textbf{Model} & \textbf{Precision} & \textbf{Recall} & \textbf{F\textsubscript{1}-Score} \\ \hline
\textbf{\begin{tabular}[c]{@{}l@{}}Proposed Model\\ GeM(RoBERTa+EM+A+TA)\end{tabular}} & \textbf{84.18} & \textbf{85.29} & \textbf{84.73} \\ \hline
\textbf{-Attention (A)} & 81.22 (2.96$\downarrow$) & 82.13 (3.16$\downarrow$ ) & 81.67 (3.06$\downarrow$) \\ \hline
\textbf{-Entity Masking (EM)} & 79.34 (4.84$\downarrow$) & 80.68 (4.61$\downarrow$) & 80.00 (4.73$\downarrow$) \\ \hline
\textbf{-Task Adaptation (TA)} & 83.25 (0.93$\downarrow$) & 84.13(1.16$\downarrow$) & 83.68 (1.05$\downarrow$) \\ \hline
\end{tabular}
}
\caption{Ablation Study: Median of metrics over 10 different runs. Bold denotes best performance.}
\label{ablation}
\vspace{-5mm}
\end{table}

\section{Conclusion}
In this paper we explored a new health dimension of social media in CVD to identify MH symptoms. We created a domain-specific dataset (CVDS) for identifying MH symptoms with underlying gender. We developed a novel framework and a computational method \model\ for understanding gender language in MH-based text and asynchronous conversations on MH platforms. We introduced a task adaptive RoBERTa-based attention framework that learns to identify the MH symptoms with the underlying task of gender detection. Our computational approach effectively identified gender while learning the MH symptoms in the CVD context. Our experimental results show the effectiveness of studying gender-based MH symptoms in a specific disease context. In the future, we aim to create a gold standard dataset with expert annotations and test our model for EMR data to better understand the gains and pitfalls in actual clinical settings. Through GeM, we hope to form a future component to process a larger EMR data set to (a) validating the new clinical process guidelines for patients with CVD and creating gender and sex-based Order Sets and updated questionnaires in EMR and (b) creating actionable gender-based order Sets integrated into the EMR system that will help the providers with decision-making and avoid possible delays and discrepancies in the Diagnosis and Treatment of CVD.
\bibliography{anthology,custom}
\bibliographystyle{acl_natbib}

\end{document}